\documentclass[runningheads]{llncs}
\usepackage[utf8]{inputenc}
\usepackage[T1]{fontenc}

\usepackage[american]{babel}

\usepackage[sort&compress,numbers]{natbib}
\makeatletter
\renewcommand\@biblabel[1]{#1.}
\makeatother
\usepackage{float}
\usepackage{mathtools}
\usepackage{amssymb}
\usepackage{bm}
\usepackage{dsfont}
\usepackage{nicefrac}
\usepackage{siunitx}
\usepackage{booktabs}
\usepackage{tabu}
\usepackage{csvsimple}
\usepackage{multicol}
\usepackage{multirow}
\usepackage[inline]{enumitem}
\usepackage{xstring}
\usepackage{tikz}
\usetikzlibrary{backgrounds,calc}
\usepackage{pgfplots}
\pgfplotsset{compat=1.5}
\pgfplotsset{
    cycle list={t_blue\\t_red\\t_green\\t_yellow\\},
}
\usepgfplotslibrary{fillbetween}
\usepackage{intcalc}
\usepackage{graphicx}
\usepackage[misc]{ifsym}
\usepackage{placeins}

\usepackage{hyperref}
\hypersetup{
    plainpages=false,           
    pdfborder={0 0 0},          
    breaklinks=true,            
    bookmarksnumbered=true,     %
    bookmarksopen=true,         %
	hypertexnames=true,         %
}

\usepackage[capitalise,nameinlink]{cleveref}
\crefname{ax}{axiom}{axioms}
\usepackage{acro}
\acsetup{format/first-long=\slshape} 
\acsetup{barriers/use=true}
\acsetup{barriers/reset=true}
\acsetup{single}

\usepackage{colortbl}
\definecolor{t_gray}{HTML}{888888}
\definecolor{t_blue}{HTML}{355fb3}
\definecolor{t_red}{HTML}{b33535}
\definecolor{t_green}{HTML}{3bb335}
\definecolor{t_yellow}{HTML}{b39735}
\definecolor{t_darkgray}{HTML}{454545}
\definecolor{t_darkblue}{HTML}{1e3666}
\definecolor{t_darkgreen}{HTML}{22661e}
\definecolor{t_darkred}{HTML}{661e1e}
\definecolor{t_darkyellow}{HTML}{66571e}
\definecolor{t_lightblue}{HTML}{8ea7d7}
\definecolor{t_lightred}{HTML}{dc8989}
\definecolor{t_lightgreen}{HTML}{8ddc89}

\pgfplotsset{
    cycle list={t_blue\\t_red\\t_green\\t_yellow\\},
}
\pgfplotscreateplotcyclelist{plotColors}{%
t_green\\%
t_blue\\%
t_red\\%
t_yellow\\%
}

\newcommand{\dac}[3]{\DeclareAcronym{#1}{short = #2, long = #3}}

\DeclareMathOperator*{\argmin}{arg\,min}

\newcommand{\bftab}{\fontseries{b}\selectfont}
\newcommand*{\condbold}[3][]{\ifthenelse{\equal{#2}{1}#1}{{\bftab #3}}{#3}}
\newcommand*{\condul}[3][]{\ifthenelse{\equal{#2}{1}#1}{{\underline{#3}}}{#3}}
\newcommand*{\tblres}[3]{\ifthenelse{\equal{#1}{}}{-}{\condbold{#3}{#1}}}

\newcommand{\approptoinn}[2]{\mathrel{\vcenter{
  \offinterlineskip\halign{\hfil$##$\cr
    #1\propto\cr\noalign{\kern2pt}#1\sim\cr\noalign{\kern-2pt}}}}}

\dac{ml}{ML}{machine learning}
\dac{ql}{QL}{quantification learning}
\dac{gql}{GQL}{graph quantification learning}
\dac{nlp}{NLP}{natural language processing}
\dac{nn}{NN}{neural network}
\dac{mlp}{MLP}{Multilayer Perceptron}
\dac{rnn}{RNN}{recurrent neural network}
\dac{svm}{SVM}{support vector machine}
\dac{cnn}{CNN}{convolutional neural network}
\dac{gnn}{GNN}{graph neural network}
\dac{gcn}{GCN}{Graph Convolutional Network}
\dac{gin}{GIN}{Graph Isomorphism Network}
\dac{gat}{GAT}{Graph Attention Network}
\dac{appnp}{APPNP}{approximate personalized propagation of neural predictions}
\dac{ce}{CE}{cross-entropy}
\dac{uce}{UCE}{uncertain cross-entropy}
\dac{ppr}{PPR}{personalized page-rank}
\dac{arc}{ARC}{accuracy-rejection curve}
\dac{ood}{OOD}{out-of-distribution}
\dac{id}{ID}{in-distribution}
\dac{auroc}{AUC-ROC}{area under the receiver operating characteristic curve}
\dac{qgnn}{QGNN}{Quantification Graph Neural Network}
\dac{cc}{CC}{Classify \& Count}
\dac{pcc}{PCC}{Probabilistic Classify \& Count}
\dac{acc}{ACC}{Adjusted Classify \& Count}
\dac{pacc}{PACC}{Probabilistic Adjusted Classify \& Count}
\dac{pps}{PPS}{prior probability shift}
\dac{mlpe}{MLPE}{Maximum Likelihood Prevalence Estimation}
\dac{slsqp}{SLSQP}{Sequential Least Squares Quadratic Programming}
\dac{bfs}{BFS}{breadth-first search}
\dac{sp}{SP}{shortest path}
\dac{rw}{RW}{random walk}
\dac{sis}{SIS}{structural importance sampling}
\dac{nacc}{NACC}{Neighborhood-aware ACC}
\dac{ae}{AE}{absolute error}
\dac{rae}{RAE}{relative absolute error}
\dac{kld}{KLD}{Kullback-Leibler divergence}
\dac{dm}{DM}{distribution matching}
\dac{rv}{RV}{random variable}
\dac{qmm}{MM}{Mixture Models}
\dac{cdf}{CDF}{cumulative distribution function}
\dac{pdf}{PDF}{probability density function}
\dac{hd}{HD}{Hellinger distance}
\dac{kde}{KDE}{kernel density estimation}
\dac{gmm}{GMM}{Gaussian mixture model}

\newcommand{\corr}{(\Letter)}

\begin{document}

\title{Distribution Matching for Graph Quantification under Structural Covariate Shift}
\titlerunning{DM for Graph Quantification under Structural Covariate Shift}
\toctitle{Distribution Matching for Graph Quantification under Structural Covariate Shift}

\author{Clemens Damke\inst{1} \corr
    \and
    Eyke Hüllermeier\inst{1,2,3}}
\tocauthor{Clemens~Damke,Eyke~Hüllermeier}

\authorrunning{C. Damke and E. Hüllermeier}

\institute{}
\institute{Institute of Informatics, LMU Munich, Germany \email{\{clemens.damke,eyke\}@ifi.lmu.de}
    \and
    Munich Center for Machine Learning (MCML)
    \and
    German Centre for Artificial Intelligence (DFKI, DSA)}

\maketitle              

\begin{abstract}
    Graphs are commonly used in machine learning to model relationships between instances.
    Consider the task of predicting the political preferences of users in a social network; to solve this task one should consider, both, the features of each individual user \emph{and} the relationships between them.
    However, oftentimes one is not interested in the label of a single instance but rather in the distribution of labels over a set of instances; e.g., when predicting the political preferences of users, the overall prevalence of a given opinion might be of higher interest than the opinion of a specific person.
    This label prevalence estimation task is commonly referred to as \ac{ql}.
    Current \ac{ql} methods for tabular data are typically based on the so-called \ac{pps} assumption which states that the label-conditional instance distributions should remain equal across the training and test data.
    In the graph setting, \ac{pps} generally does not hold if the shift between training and test data is structural, i.e., if the training data comes from a different region of the graph than the test data.
    To address such structural shifts, an importance sampling variant of the popular adjusted count quantification approach has previously been proposed.
    In this work, we extend the idea of structural importance sampling to the state-of-the-art KDEy quantification approach.
    We show that our proposed method adapts to structural shifts and outperforms standard quantification approaches.

    \keywords{Quantification Learning  \and Graph Quantification \and Covariate Shift.}
\end{abstract}
\acbarrier

\section{Introduction}\label{sec:intro}

\Ac{ql} refers to the task of estimating the distribution of labels $\mathcal{Y}$ over a set of instances $\mathcal{X}$~\citep{forman2005,forman2006,forman2008}.
More specifically, one is given a set of labeled training instances $\mathcal{D}_L \subseteq \mathcal{X} \times \mathcal{Y}$ drawn from a distribution $P$ and a set of unlabeled test instances $\mathcal{X}_U \subseteq \mathcal{X}$ drawn from a distribution $Q$ for which the label distribution $Q(Y)$ is to be estimated.
For example, the problem of predicting the prevalence of different opinions in a given population of people can be seen as a \ac{ql} task.
Here, the training data consists of a sample of people with known opinions, while the test data consists of a second sample of people with unknown opinions for which the opinion prevalences should be estimated.

A na\"ive way to solve a quantification problem is to use a standard classification model to label all test instances $\mathcal{X}_U$.
The relative frequencies of the predicted labels could then be used as an estimate of $Q(Y)$.
Given a perfect classifier, this so-called \ac{cc} strategy would indeed yield a perfect estimate of the test label distribution.
This is, however, an unrealistic assumption, leading to the question of whether an imperfect classifier can still provide good quantification results.
\Citet{forman2005} showed that the simple \ac{cc} approach can lead to poor quantification results if the classifier is biased.
To understand why, note that the goal of a classifier is to minimize the number of classification errors, i.e., the sum of false positive and false negative predictions ($\mathrm{FP} + \mathrm{FN}$) in the binary case.
In contrast, the goal of a quantifier is to minimize $\left|\mathrm{FP} - \mathrm{FN}\right|$; if $\mathrm{FP} = \mathrm{FN}$, the missing positive predictions are perfectly compensated for by the missing negative predictions, leading to a perfect quantification result.
This implies that even a poor classifier (in terms of misclassifications) can provide good quantification results and vice versa.
Therefore, quantification should be treated as a distinct task from classification~\citep{esuli2023}.

Suppose the training and test data are drawn from the same distribution ($P = Q$).
In that case, the quantification task is trivially solved using the label distribution of the sampled training data $\mathcal{D}_L$ as an unbiased estimate of the label distribution of the test data.
The \ac{ql} problem gets more challenging when the training and test data are drawn from different distributions, i.e., in the presence of so-called \emph{distribution shift}.
In this case, the distribution of training labels $P(Y)$ is not necessarily a good estimate of the test label distribution $Q(Y)$.
In the extreme case, if the two sampling distributions are entirely unrelated and the training data thus is uninformative about the test data, the quantification task becomes intractable without other prior assumptions.

Therefore, one typically assumes at least some kind of relation between the training and test data.
The most commonly assumed type of shift is called \ac{pps}~\citep{moreno-torres2012,gonzalez2024}, which states that the class-conditional instance distributions remain unchanged between $P$ and $Q$.
This assumption is made, among others, by the well-known \ac{acc} quantification method~\citep{forman2005} and by \ac{dm} methods, such as the \ac{qmm} approach~\citep{forman2005}, HDy~\citep{gonzalez-castro2013}, DyS~\citep{maletzke2019} or KDEy~\citep{moreo2025}.

While \ac{pps} is a reasonable assumption in many domains, there are problems where it does not hold.
If the training and test data are drawn from different regions of the instance space, the \ac{pps} assumption may not be satisfied.
In this case, $P$ and $Q$ may instead be related by \emph{covariate shift}~\citep{moreno-torres2012}, which states that the distributions of the instances can differ, while the instance-conditional label distributions remain unchanged.
\Citet{gonzalez2024} and \citet{tasche2022} show, both empirically and theoretically, that the standard \ac{acc} and \ac{dm} approaches are not robust to covariate shift.

One domain where covariate shift arises naturally is \emph{node label quantification} in graph data.
Here, instances are nodes in a graph which are connected by edges indicating some notion of relatedness.
Consider the opinion quantification problem mentioned earlier, where people can be naturally represented as nodes in a social network with edges indicating social relations (coworkers, friendships, or familial relations).
The structural information contained in such graph representations has been used with great success by \ac{gnn} models to solve node classification tasks~\citep{khemani2024}.
However, there has been little work on \ac{gql}.
\Citet{milli2015} and \citet{tang2010} have proposed simple \ac{gql} methods based on community detection algorithms which do not make use of current predictive graph models.
Recently, \citet{damke2025} proposed the first classifier-based quantification method for graph data.
They extend \ac{acc} to account for structural covariate shift by introducing the kernel-based \ac{sis} method.
\Ac{acc} with \ac{sis} is shown to give unbiased estimates of $Q(Y)$ under covariate shift.
However, state-of-the-art \ac{dm} methods, such as KDEy, generally tend to outperform \ac{acc} methods under \ac{pps}~\citep{moreo2025}.
Translating the practical advantages of \ac{dm} to the covariate shift setting via \ac{sis} is therefore desirable.

In this work, we extend \ac{sis} to the \ac{dm} framework and adapt KDEy to structural covariate shift and show that this combination outperforms previously proposed quantification approaches.
To this end, we begin with a brief introduction to \ac{ql} in general and \ac{dm} methods in particular (\cref{sec:ql}).
\Cref{sec:approach} then describes the \ac{sis} method and how it can be used in the \ac{dm} framework.
In \cref{sec:eval}, we evaluate the proposed method on a set of benchmark datasets under different types of distribution shift and compare it to other quantification approaches.
Finally, we conclude with a brief outlook in \cref{sec:conclusion}.

\section{An Overview of Quantification Learning}\label{sec:ql}

In the literature on \ac{ql}, one typically distinguishes between \emph{aggregative} and \emph{non-aggregative} approaches~\citep{esuli2023}.
Aggregative methods are based on a standard classification model that is trained on labeled training instances.
The predictions of this model on the test data are aggregated to estimate the test label distribution.
Non-aggregative methods, in contrast, are directly trained to predict label prevalences given a set of instances.
Here, we focus on aggregative methods; the extension of non-aggregative quantification methods to covariate shift is left for future work.
First, we fix some notation and formally define the \acl{ql} setting.

\subsection{Notation and Problem Setting}\label{sec:ql:notation}

As described in the introduction, let $\mathcal{X}$ denote the instance space and $\mathcal{Y} = \{1, \dots, K\}$ the (finite) label space.
Let $P$ and $Q$ be probability measures on $\mathcal{X} \times \mathcal{Y}$ representing the training and test data distributions, respectively.
Let $X$ and $Y$ denote \acp{rv} that project the joint instance-label space to the instance and label spaces, respectively.
In \ac{ql}, we are given a labeled sample $\mathcal{D}_L \subseteq \mathcal{X} \times \mathcal{Y}$ drawn from $P$ and an unlabeled sample $\mathcal{X}_U \subseteq \mathcal{X}$ drawn from $Q(X) = Q \circ X^{-1}$.
The goal of \ac{ql} is to estimate $Q(Y)$ from $\mathcal{D}_L$ and $\mathcal{X}_U$.
The two measures $P$ and $Q$ are assumed to be related by some kind of distribution shift~\citep{moreno-torres2012}:
\begin{enumerate}[label=(\roman*)]
    \item
          In \acf{pps}, $P(Y)$ and $Q(Y)$ might differ but the label-conditional instance distributions remain equal, i.e., $P(X \mid Y) = Q(X \mid Y)$.
    \item
          In \emph{covariate shift}, the distributions of the instances $P(X)$ and $Q(X)$ differ, while the conditional label distributions remain unchanged, i.e., $P(Y \mid X) = Q(Y \mid X)$.
    \item
          In \emph{concept shift}, it is the conditional label distributions $P(Y \mid X)$ and $Q(Y \mid X)$ that change while the marginal instance distributions $P(X)$ and $Q(X)$ remain equal.
\end{enumerate}
\begin{figure}[t]
    \centering
    \includegraphics[width=0.7\linewidth]{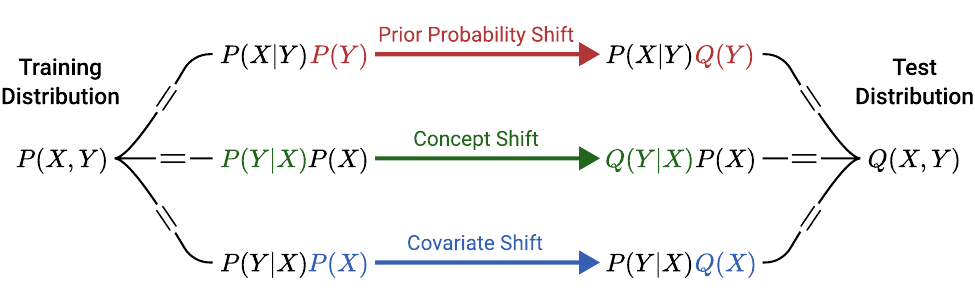}
    \caption{Overview of the three typically considered types of distribution shift.}\label{fig:shift-types}
\end{figure}
See \cref{fig:shift-types} for an overview.
Depending on the problem domain, different types of distribution shift can occur~\citep{moreno-torres2012}.
For example, \ac{pps} might arise in the context of epidemiological studies where the task is to estimate the prevalence of a disease in a population.
Here, the training data might be collected via a case-control study where the percentage of healthy and infected people is fixed by design, while the test data is collected from a random sample of the population.
In contrast, the opinion estimation problem mentioned earlier might be subject to covariate shift if the training data is collected in a different region of the population than the test data.
Last, concept shift occurs if the meaning of labels change between training and test data, e.g., whether a newspaper article is about a local or world news depends on the location of the reader/newspaper.

Assuming \ac{pps}, the training and test distributions are related by $P(X \mid Y) = Q(X \mid Y)$.
Consequently, for any measurable mapping $\phi: \mathcal{X} \to \mathcal{Z}$, we have $P(Z \mid Y) = Q(Z \mid Y)$ where $Z = \phi(X)$~\citep{lipton2018}.
This allows us to factorize $Q(Z)$ as follows:
\begin{align}
    Q(Z) = \sum_{i=1}^K Q(Z \mid Y = i) Q(Y = i) = \sum_{i=1}^K P(Z \mid Y = i) Q(Y = i) \label{eq:pps-factorization}
\end{align}
Given $\mathcal{D}_L$ and $\mathcal{X}_U$, both, $Q(Z)$ and $P(Z \mid Y)$ can (in principle) be estimated, resulting in a system of equations which can be solved for the desired test label distribution $Q(Y)$.
This idea forms the basis for many \ac{ql} methods, including \ac{acc}~\citep{forman2005,forman2008,bunse2022} and the family of \ac{dm} methods~\citep{firat2016,bunse2022a,moreo2025}.
The main difference between those methods lies in how the mapping $\phi$ is chosen and how the system of equations is solved.
One common approach is to define $\phi$ in terms of a hard classifier $h: \mathcal{X} \to \mathcal{Y}$ or a probabilistic classifier $h_s: \mathcal{X} \to \Delta_K$, where $\Delta_K$ is the unit $(K-1)$-simplex.
Let $\hat{Y} = h(X)$ denote the predicted label and $\hat{S}_i = h_s(X)_i$ be the predicted probability of label $i$.

For $\phi = h$, \cref{eq:pps-factorization} turns into \ac{acc}, where $P(\hat{Y} \mid Y)$ is simply the confusion matrix of $h$.
\Citet{tasche2017} shows that this results in an unbiased estimate of $Q(Y)$ under \ac{pps}.
Similarly, using a confusion matrix derived from $h_s$ leads to the \ac{pacc} approach~\citep{bella2010}, while the unadjusted pendant of \ac{cc} is referred to as \ac{pcc}.
Next, we will describe the \acl{dm} framework.


\subsection{Histogram-based DM Approaches}\label{sec:ql:histogram}

The first \ac{dm} quantification method, simply refered to as \acf{qmm}, was proposed by \citet{forman2005}.
\Ac{qmm} is designed for binary quantification problems, i.e., $\mathcal{Y} = \{\oplus, \ominus\}$, and uses $\phi = h_s$, where $h_s: \mathcal{X} \to [0,1]$ is a soft binary classifier.
\Cref{eq:pps-factorization} then becomes
\begin{align*}
    Q(\hat{S}) = {\underbrace{P(\hat{S} \mid Y = \oplus)}_{P_\oplus} \underbrace{Q(Y = \oplus)}_{\alpha}} + {\underbrace{P(\hat{S} \mid Y = \ominus)}_{P_\ominus} \underbrace{Q(Y = \ominus)}_{1 - \alpha}} \ .
\end{align*}
Here, the distribution of predicted probabilities $Q(\hat{S})$ is modeled as a mixture of the class-conditional predicted probability distributions $P_\oplus, P_\ominus$.
\Citeauthor{forman2005} estimates those distributions via discrete \acp{cdf} $\hat{p}_\oplus, \hat{p}_\ominus$ predicted by $h_s$ on $\mathcal{D}_L$.
Analogously, an estimate $\hat{q}$ of $Q(\hat{S})$ is obtained by computing the discrete \ac{cdf} using $\mathcal{X}_U$.
The mixture weight $\alpha$ is then determined by solving the following optimization problem:
\begin{align}
    \alpha^* = \argmin_{\alpha \in [0,1]} \ell\left( \alpha \cdot \hat{p}_\oplus + (1-\alpha) \cdot \hat{p}_\ominus, \hat{q} \right)\ , \label{eq:mm-optimization}
\end{align}
where $\ell$ is a loss function measuring the discrepancy between the mixture \ac{cdf} and $\hat{q}$.
\Citeauthor{forman2005} proposes two different loss functions: PP-Area, which is equivalent to minimizing the L1-norm between the two \acp{cdf}~\citep{firat2016}, and the Kolmogorov-Smirnov statistic.
\citet{gonzalez-castro2013} extend \ac{qmm} by proposing a variant that estimates the \acp{pdf} of $P_\oplus$, $P_\ominus$ and $Q(\hat{S})$ via normalized histograms $\hat{p}_\oplus, \hat{p}_\ominus, \hat{q} \in \mathbb{R}^b$ instead of discrete \acp{cdf} estimates; here, $b \in \mathbb{N}_0$ is the number of bins.
Additionally, they suggest using the \ac{hd} as a loss function $\ell$ between \acp{pdf}:
\begin{align}
    \mathrm{HD}(p, q) & \coloneqq \frac{1}{\sqrt{2}} \|\sqrt{p} - \sqrt{q}\|_2 \label{eq:hd}
\end{align}
This variant of \ac{qmm} is referred to as HDy.
Note that, both, \ac{qmm} and HDy can only be applied to binary quantification problems.
Extensions of HDy to the multi-class regime have been proposed by \citet{firat2016} and \citet{bunse2022a}.
If $K > 2$, one can compute a class-conditional histogram $\hat{p}_{j,i} \in \mathbb{R}^b$ of $P(\hat{S}_j \mid Y = i)$ for each pair of classes $j,i$ and one histogram $\hat{q}_j$ of $Q(\hat{S}_j)$ for each $i \in \mathcal{Y}$.
To obtain a representation of $P(\hat{S} \mid Y = i)$, one can then combine the class-conditional histograms by concatenating and renormalizing them to obtain a single histogram $\hat{p}_i = \frac{1}{K} \left(\hat{p}_{1,i},\dots,\hat{p}_{K,i}\right) \in \mathbb{R}^{bK}$~\citep{firat2016}, or by directly averaging them, i.e., $\hat{p}_i = \frac{1}{K}\sum_{j=1}^K \hat{p}_{j,i}$~\citep{bunse2022a}.
\Citet{moreo2025} show that both of these histogram aggregation variants are theoretically flawed, since neither of them produces a proper estimate of the divergence between \acp{pdf}.
The problem of those multi-class extensions of HDy is that the per-class histograms $\hat{p}_{j,i}$ and $\hat{q}_j$ are unable to capture inter-class information.

\subsection{Kernel Density Estimation-based DM}\label{sec:ql:kdey}

To address the shortcomings of histogram-based \ac{dm} methods, \citet{moreo2025} propose the KDEy quantification approach.
Unlike \ac{qmm} and HDy, KDEy does not represent $Q(\hat{S})$ and $P(\hat{S} \mid Y)$ by decomposing them into class-wise histograms but instead uses \ac{kde} to model the \acp{pdf} of the predicted probabilities as \acp{gmm}.
More specifically, let $q(\hat{s})$ be the \ac{pdf} of $Q(\hat{S})$ and $p(\hat{s} \mid i)$ be the \ac{pdf} of $P(\hat{S} \mid Y = i)$, with $s = (s_1, \dots, s_K) \in \Delta_{K}$ being a vector of predicted label probabilities.
Using \ac{kde}, we can estimate those \acp{pdf} as follows:
\begin{align}
    \hat{q}(\hat{s}) = \frac{1}{|\mathcal{X}_U|} \smashoperator[r]{\sum_{x \in \mathcal{X}_U}} k(\hat{s}, h_s(x)) \quad \text{and} \quad \hat{p}(\hat{s} \mid i) = \frac{1}{|\mathcal{D}_L^i|} \smashoperator[r]{\sum_{(x,y) \in \mathcal{D}_L^i}} k(\hat{s}, h_s(x))\ , \label{eq:kdey}
\end{align}
where $k: \Delta_K \times \Delta_K \to \mathbb{R}_{\geq 0}$ is a kernel function and $\mathcal{D}_L^i \coloneqq \{(x,y) \in \mathcal{D}_L \mid y = i\}$.
In KDEy, the kernel $k$ is chosen to be a Gaussian kernel with bandwidth $\sigma$:
\begin{align}
    k(\hat{s}, \hat{s}') \coloneqq \frac{1}{\sqrt{(2\pi)^K} \sigma^K} \exp\left(-\frac{1}{2\sigma^2} \|\hat{s} - \hat{s}'\|_2^2\right)\ .
\end{align}
The bandwidth $\sigma$ is treated as a model hyperparameter.
Using the linearity of the Radon-Nikodym derivative\footnote{
    Note that $q(\hat{s}) = \frac{d Q(\hat{S})}{d \mu}$ and $p(\hat{s} \mid i) = \frac{d P(\hat{S} \mid Y = i)}{d \mu}$, with the Lebesgue measure $\mu$. 
} we can plug our \ac{pdf} estimates into \cref{eq:pps-factorization} to obtain the following equation:
\begin{align}
    \hat{q}(\hat{s}) = \sum_{i=1}^K \hat{p}(\hat{s} \mid i) q(i)\ , \label{eq:kdey-factorization}
\end{align}
where both sides of the equation are \acp{gmm} and $q \in \Delta_K$ is the vector of label prevalences in the test distribution which we are looking for.
Analogous to the histogram-based \ac{dm} approaches, we can now solve for $q$ by minimizing a divergence $\ell$ between the left-hand side and the right-hand side of \cref{eq:kdey-factorization}:
\begin{align}
    \hat{q} = \argmin_{q \in \Delta_K} \ell\left( \hat{q}(\hat{s}), \sum_{i=1}^K q_i \cdot \hat{p}(\hat{s} \mid i)\right)\ . \label{eq:kdey-optimization}
\end{align}
Different choices of $\ell$ are possible here, e.g., \ac{hd} (see \cref{eq:hd}), L2, Cauchy-Schwarz, Jensen-Shannon or \ac{kld}.
Since $\ell$ has to be computed for continuous distributions which, depending on the divergence, can be computationally intractable, \citet{moreo2025} suggest a number of divergence-dependent optimization strategies.
For the Cauchy-Schwarz divergence, they derive a closed-form solution to \cref{eq:kdey-optimization}.
For the \ac{hd}, Jensen-Shannon and L2 divergences, they propose a Monte Carlo approximation approach.
For the \ac{kld}, they show that \cref{eq:kdey-optimization} reduces to
\begin{align}
    \hat{q} = \argmin_{q \in \Delta_K} - \smashoperator{\sum_{x \in \mathcal{X}_U}} \log \sum_{i=1}^K q_i \cdot \hat{p}(h_s(x) \mid i) \ , \label{eq:kdey-kld}
\end{align}
which can be solved using standard (gradient-based) constrained optimization techniques.
In their experiments, the different variants of KDEy generally outperform the histogram-based \ac{dm} methods, with the \ac{hd}- and \ac{kld}-based variants performing particularly strongly.

\section{DM under Structural Covariate Shift}\label{sec:approach}

The \ac{dm} approaches for quantification described in the previous section are all based on the \ac{pps} assumption.
However, as discussed before, this assumption is not always justified in practice.
When dealing with graph data, covariate shifts are of particular interest.
If the training data is collected from a different region of the graph than the test data, there is so-called \emph{structural covariate shift} between $P$ and $Q$.
\Citet{damke2025} propose an extension of \ac{acc} to account for such structural covariate shifts via so-called \acf{sis}.
We will now show how this idea can be extended to the \ac{dm} framework.

\begin{figure}[t]
    \centering
    \includegraphics[width=\linewidth]{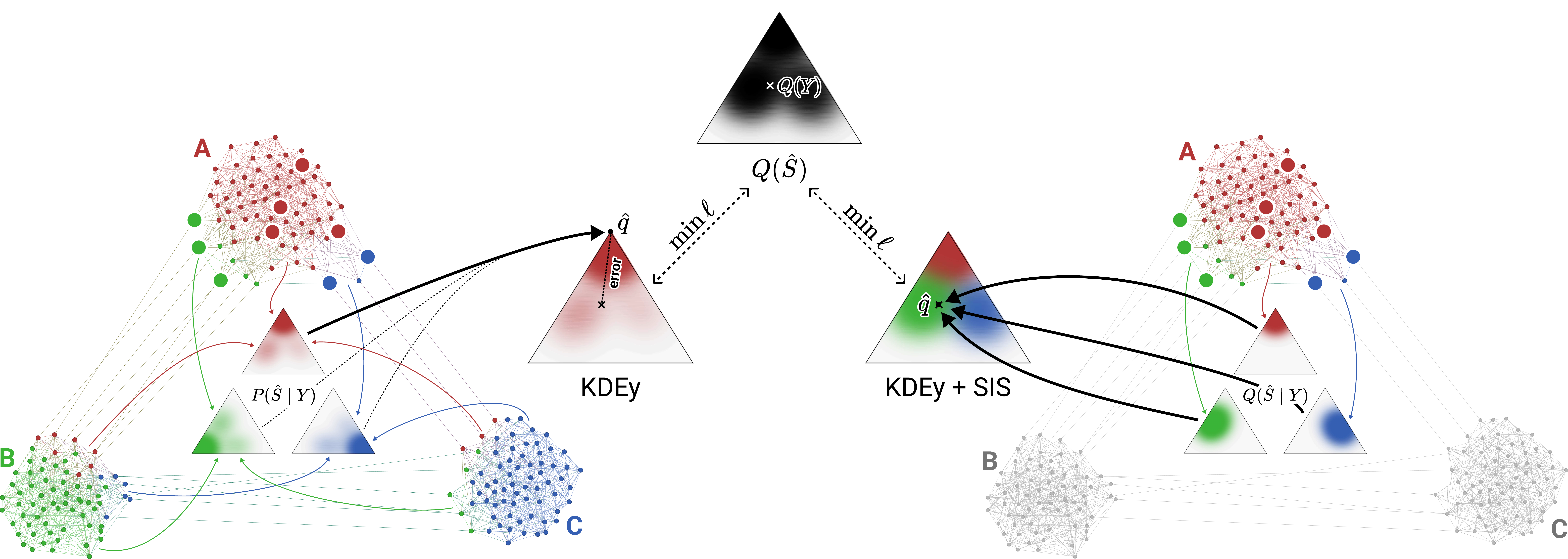}
    \caption{Illustration of the advantage of \ac{sis} in the \ac{dm} framework under structural covariate shift.}\label{fig:kdey-sis}
\end{figure}

To motivate our approach, consider the following example:
\def\clA{\textcolor{t_red}{\textsf{\textbf{A}}}}
\def\clB{\textcolor{t_green}{\textsf{\textbf{B}}}}
\def\clC{\textcolor{t_blue}{\textsf{\textbf{C}}}}
\Cref{fig:kdey-sis} shows a simple graph consisting of three vertex clusters, each corresponding to one label $\mathcal{Y} = \{\clA, \clB, \clC\}$.
Most vertices in each cluster have the matching label; there are, however, a few outliers incident to the edges between clusters.
Assume that the training data $\mathcal{D}_L$ is sampled uniformly at random from all three clusters and that the test data $\mathcal{X}_U$ comes only from the topmost cluster \clA\ (indicated by the large nodes), i.e., there is covariate shift between both samples.
A probabilistic classifier $h_s$ has high confidence on inlier vertices and low confidence for outliers.
The simplex plots on the left of \cref{fig:kdey-sis} show the \acp{pdf} of $P(\hat{S} \mid Y)$ for all three labels $\mathcal{Y}$.
The standard KDEy method described in \cref{sec:ql:kdey} tries to find an optimal mixture of those \acp{pdf} to match the target \ac{pdf} $Q(\hat{S})$ of the test data (shown in black) by minimizing a divergence $\ell$ between the mixture \ac{pdf} and $Q(\hat{S})$.
Since outliers are rare for each class, the \acp{pdf} $P(\hat{S} \mid Y)$ are concentrated around the high confidence regions of each class.
$\mathcal{X}_U$, on the other hand, contains many outliers, causing the test \ac{pdf} $Q(\hat{S})$ to be less concentrated.
In this case, the optimal mixture found by KDEy will put all weight on $P(\hat{S} \mid Y = \clA)$, resulting in a heavily biased estimate of $Q(Y)$.

The problem is that the global \acp{pdf} $P(\hat{S} \mid Y)$ are not representative of the local (shifted) \acp{pdf} $Q(\hat{S} \mid Y)$.
Instances with label \clB\ and \clC\ within cluster \clA\ will receive low confidence scores, as shown in the simplex plots on the right side of \cref{fig:kdey-sis}.
By combining these estimates of $Q(\hat{S} \mid Y)$, as opposed to $P(\hat{S} \mid Y)$, one can find a \ac{pdf} mixture that better matches $Q(\hat{Y})$ resulting in a better predicted label distribution.
The core idea behind \ac{sis} is to use the graph structure to obtain such estimates.
We will now describe \ac{sis} more formally and then adapt it to the \ac{dm} framework.

\subsection{Structural Importance Sampling for ACC}\label{sec:approach:sis}

As described in \cref{sec:ql:notation}, both, \ac{acc} and \ac{dm} methods are based on the factorization in \cref{eq:pps-factorization}.
Under covariate shift, this factorization does not hold, since $Q(Z \mid Y)$ is unknown.
If $\phi = h$, i.e., $Z = \hat{Y} = h(X)$, we can rewrite $Q(\hat{Y} \mid Y)$ via importance sampling~\citep{damke2025}:
\begin{align}
    Q(\hat{Y} = j \mid Y = i) & = \int_{x\in\mathcal{X}} \mathds{1}[h(x) = j] d Q(X = x \mid Y = i) \nonumber                                                                                            \\
                              & = \int_{x\in\mathcal{X}} \mathds{1}[h(x) = j] \underbrace{\frac{q_{X|Y}(x \mid i)}{p_{X|Y}(x \mid i)}}_{= \rho_{X|Y}(x \mid i)} d P(X = x \mid Y = i) \label{eq:sis-acc}
\end{align}
Here, $q_{X|Y}$ and $p_{X|Y}$ denote the conditional \acp{pdf} of $Q$ and $P$, and $\rho_{X|Y}$ denotes the ratio between those densities.
Under (structural) covariate shift, we know that $p_{Y|X} = q_{Y|X}$, which allows us to rewrite $\rho_{X|Y}$:
\begin{align}
    \rho_{X|Y}(x \mid y) & = \frac{q_{X|Y}(x \mid y)}{p_{X|Y}(x \mid y)} = \frac{q_{Y|X}(y \mid x) q_X(x) p_Y(y)}{p_{Y|X}(y \mid x) p_X(x) q_Y(y)} = \rho_X(x) \cdot \rho_Y(y)^{-1} \label{eq:sis-rho}
\end{align}
Plugging this into \cref{eq:sis-acc} gives us
\begin{align}
    Q(\hat{Y} = j \mid Y = i) 
     & = \frac{\rho_Y(i)^{-1} \int_{\mathcal{X}} \mathds{1}[h(x) = j] \rho_X(x) d P(X = x \mid Y = i)}{\rho_Y(i)^{-1} \int_{\mathcal{X}} \rho_X(x) d P(X = x \mid Y = i)} \nonumber \\
     & = \frac{\mathbb{E}_{P(X \mid Y=i)}[\mathds{1}[\hat{Y} = j] \rho_X(X)]}{\mathbb{E}_{P(X \mid Y=i)}[\rho_X(X)]}\ .
\end{align}
To compute $\rho_X = \frac{q_X}{p_X}$, \citet{damke2025} propose to use \acl{kde}.
Since we are given samples $\mathcal{D}_L$ from $P$ and samples $\mathcal{X}_U$ from $Q(X)$, $q_X$ and $p_X$ can be estimated as follows:
\begin{align}
    \hat{q}_X(x) & = \frac{1}{|\mathcal{X}_U|} \smashoperator[r]{\sum_{x' \in \mathcal{X}_U}} \kappa(x, x') \quad \text{and} \quad \hat{p}_X(x) = \frac{1}{|\mathcal{D}_L|} \smashoperator[r]{\sum_{(x',y') \in \mathcal{D}_L}} \kappa(x, x')\ , \label{eq:sis-kde}
\end{align}
where $\kappa: \mathcal{X} \times \mathcal{X} \to \mathbb{R}_{\geq 0}$ is a suitable instance kernel function.
Without any domain assumptions, choosing $\kappa$ appropriately is difficult.

However, in the context of graph data where instances $x$ are vertices, assuming structural covariate shift, the probability of sampling a vertex $x$ from $P$ or $Q$ should depend on how close $x$ is to the training or test vertices, respectively.
\Citeauthor{damke2025} suggest that the appropriate notion of ``closeness'' in a graph depends on the nature of the covariate shift.
For example, if the data is sampled via random walks, a \ac{ppr} kernel~\citep{page1999} is appropriate:
\begin{align}
    \kappa_{\mathrm{PPR}}(x_i, x_j) = \Pi_{i,j}, \quad \text{where} \quad \Pi = {\left(\alpha\mathbf{I} + (1-\alpha) \bar{\mathbf{A}}\right)}^L\ . \label{eq:ppr-kernel}
\end{align}
Here, $\bar{\mathbf{A}} = \mathbf{A} \mathbf{D}^{-1}$ is the normalized adjacency matrix of the graph, $\alpha \in (0,1)$ is a teleportation parameter and $L$ is the number of steps in the random walk.
If the vertex sampling process is based on the shortest path lengths between vertices, a \ac{sp} kernel can be used, e.g.,
\begin{align}
    \kappa_{\mathrm{SP}}(x_i, x_j) = \exp\left(-\lambda \cdot d_{\mathrm{SP}}(x_i, x_j )\right), \label{eq:sp-kernel}
\end{align}
where $d_{\mathrm{SP}}(x_i, x_j)$ is the length of the shortest path between $x_i$ and $x_j$ and $\lambda$ is a scaling parameter.
To summarize, $k$ should be chosen based on available knowledge about the quantification problem at hand to reflect the nature of the shift as closely as possible.

\subsection{SIS in the DM Framework}\label{sec:approach:sis-dm}

In \ac{acc}, the instance mapping $\phi$ reduces an instance $X$ to a single label prediction $\hat{Y}$, discarding much, potentially valuable, information.
As discussed in \cref{sec:ql}, \ac{dm} methods instead consider the distribution of predicted probability vectors $\hat{S} = h_s(X)$.
To apply \ac{sis} in the \ac{dm} framework, we can rewrite $q(\hat{s} \mid y)$:
\begin{align}
    q(\hat{s} \mid y) & = \int_{x\in\mathcal{X}} q(\hat{s} \mid x) d Q(X=x \mid Y=y)                                                                                   \nonumber \\
                      & =  \int_{x\in\mathcal{X}} p(\hat{s} \mid x) \underbrace{\frac{q_{X|Y}(x \mid y)}{p_{X|Y}(x \mid y)}}_{= \rho_{X|Y}(x \mid y)} d P(X=x \mid Y=y)
\end{align}
Analogous to \ac{sis} for \ac{acc}, we can use \cref{eq:sis-rho} for the following replacement:
\begin{align}
    q(\hat{s} \mid Y = i) & = \frac{\mathbb{E}_{P(X \mid Y=y)}[p(\hat{s} \mid X) \rho_X(X)]}{\mathbb{E}_{P(X \mid Y=y)}[\rho_X(X)]}\ .
\end{align}
Given $\mathcal{D}_L$ and $\mathcal{X}_U$ we can compute an estimate $\hat{\rho}_X(x)$ via \cref{eq:sis-kde}.
Additionally, $q(\hat{s} \mid Y = i)$ can be estimated via
\begin{align}
    \hat{q}(\hat{s} \mid y) & = \frac{1}{\sum_{(x,y) \in \mathcal{D}_L^i} \hat{\rho}_X(x)} \smashoperator[r]{\sum_{(x,y) \in \mathcal{D}_L^i}} p(\hat{s} \mid x) \cdot \hat{\rho}_X(x)
\end{align}
Since $p(\hat{s} \mid x) = \delta[\hat{s} = h_s(x)]$ is a Dirac delta, i.e., the density is zero everywhere except at $\hat{s} = h_s(x)$, this estimate is not particularly useful given a finite sample $\mathcal{D}_L$.
To account for this, we replace the Dirac delta with a Gaussian kernel $k: \Delta_K \times \Delta_K \to \mathbb{R}_{\geq 0}$ and obtain
\begin{align}
    \hat{q}(\hat{s} \mid y) & = \frac{1}{\sum_{(x,y) \in \mathcal{D}_L^i} \hat{\rho}_X(x)} \smashoperator[r]{\sum_{(x,y) \in \mathcal{D}_L^i}} k(\hat{s}, h_s(x)) \cdot \hat{\rho}_X(x)\ .
\end{align}
The result is a weighted version of KDEy (cf.~\cref{eq:kdey}), where the instance weights $\hat{\rho}_X(x)$ are themselves estimated via \ac{kde} using a vertex kernel $\kappa$.
By reweighting the labeled samples $\mathcal{D}_L$ via \ac{sis}, KDEy can be applied to quantification problems with structural covariate shift.

\section{Evaluation}\label{sec:eval}

We evaluate the proposed combination of \ac{sis} and KDEy on a set of benchmark datasets under different types of distribution shift using multiple node classifiers and quantification metrics.
We compare our approach against \ac{pcc}, \ac{pacc}, \ac{pacc} with \ac{sis} and standard KDEy without \ac{sis}.
We use the \texttt{QuaPy} Python library~\citep{moreo2021a} and \texttt{torch-geometric}~\citep{fey2019} to implement our experiments\footnote{Code available at \url{https://github.com/Cortys/graph-quantification}.}.
For efficient GPU-based sampling of \ac{bfs}-based test sets and the computation of the \ac{sp} kernels, we use Nvidia's \texttt{cuGraph} library.
All experiments were conducted using an AMD Ryzen 9 7950X CPU, 64GB RAM and an Nvidia RTX 4090 GPU.

\subsection{Experimental Setup}\label{sec:eval:setup}

\paragraph{Quantification Metrics}
To compare the quality of a label distribution estimate $\hat{q}$ against the ground-truth label distribution $q$, we use the following two common quantification metrics: \Acf{ae} and \acf{rae}:
\begin{align}
    \mathrm{AE}(q, \hat{q}) & = \frac{1}{K} \sum_{i=1}^K |q_i - \hat{q}_i| &
    \mathrm{RAE}(q, \hat{q}) = \frac{1}{K} \sum_{i=1}^K \frac{|q_i - \hat{q}_i|}{q_i}
\end{align}
While \ac{ae} penalized all errors equally, \ac{rae}~\citep{gonzalez-castro2013} penalizes errors on rare labels more heavily.

\paragraph{Datasets}
We generate quantification tasks from the following five node classification datasets:
\begin{enumerate*}
    \item CoraML,
    \item CiteSeer,
    \item PubMed,
    \item Amazon Photos and
    \item Amazon Computers
\end{enumerate*}~\citep{mccallum2000,giles1998,getoor2005,sen2008,namata2012,mcauley2015,shchur2019}.
The first three datasets are citation networks, where the nodes are documents and the edges represent citations between them.
The two Amazon datasets are product co-purchasing graphs, where the nodes are products and the edges represent that are often bought together.
All nodes are labeled with the topic or product category they belong to.
All datasets were split randomly 10 times into three partitions classifier train, quantifier train and quantifer test with sizes $5\%$/$15\%$/$80\%$.
Using those splits, we train each classifier 10 times on each of the 10 classifier train sets and use each of the resulting 100 classifiers per dataset with each type of quantifier.

\paragraph{Distribution Shift}
To evaulate the behavior of the quantifiers under distribution shift, we synthetically introduce shifts to the test partitions of the datasets, while the training data is sampled uniformly at random from the training split.
We consider the following types of distribution shift:
\begin{enumerate}
    \item \ac{pps}: We sample $10 \cdot K$ sets of 100 nodes such that each set has a prescribed label distribution $q \in \Delta_K$ which is sampled from a Zipf distribution over the labels~\citep{qi2020}.
    \item Structural covariate shift via \acp{rw}: For each label, we select 10 corresponding vertices and for each of those vertices we sample 100 nodes via random walks of length 10 with teleportation parameter $\alpha = 0.1$.
    \item Structural covariate shift via \ac{bfs}: Analogous to the \ac{ppr} setting, we also evaluate structural covariate shift by sampling 100 nodes via breadth-first search instead of random walks.
\end{enumerate}

\paragraph{Classifiers}
We use four types of vertex classifiers:
\begin{enumerate*}
    \item A standard \ac{mlp} which does not use any graph information,
    \item \Ac{gcn}~\citep{kipf2017},
    \item \Ac{gat}~\citep{velickovic2018} and
    \item \Ac{appnp}~\citep{gasteiger2018}.
\end{enumerate*}
All models consist of two hidden fully connected layers and two convolutional layers (where applicable) with widths of 64 and ReLU activations.

\paragraph{Quantifiers}
We compare \ac{pacc} and KDEy with and without \ac{sis}.
Additionally, we include \ac{pcc}, as it should, in principle, be able to account for covariate shift to some extent~\citep{tasche2022,gonzalez2024}.
For KDEy, we use the \ac{kld} as the divergence, referred to as KDEy-ML by \citet{moreo2025}, as this variant generally produces good quantification results while also being computationally tracktable.
For \ac{sis}, we use an interpolated version of the \ac{ppr} kernel from \cref{eq:ppr-kernel} for the \ac{kde} estimate of $q_X$:
\begin{align*}
    \kappa_{\lambda}(x, x') = \lambda \kappa_{\mathrm{PPR}}(x, x') + (1-\lambda) \ ,
\end{align*}
where $\lambda \in [0,1]$ is a hyperparameter that controls the minimum weight that should be assigned to each vertex.
For the \ac{kde} estimate of $p_X$, we use a constant kernel $\kappa_{1}(x, x') = 1$ since the training data is not subject to synthetic distribution shift in our setup.
This implies that $\rho_X = q_X$, simplifying the \ac{sis} estimation.
Additionally, we evaluate \ac{sis} with the \ac{sp} kernel from \cref{eq:sp-kernel} with $\lambda = \frac{1}{2}$ in the \acs{bfs}-based covariate shift setting to check whether the distance-based \ac{bfs} sampling is better matched by this kernel.

\subsection{Experimental Results}\label{sec:eval:results}

\newcommand{\quantTable}[1]{\renewcommand{\arraystretch}{0.865}\tiny%
    \csvreader[
        column count=70,
        tabular={cl | rr | rr | rr | rr | rr | rr},
        separator=comma,
        table head={%
                Model& &%
                \multicolumn{2}{c}{\textbf{CoraML}} &%
                \multicolumn{2}{c}{\textbf{CiteSeer}} &%
                \multicolumn{2}{c}{\textbf{A.\ Photos}} &%
                \multicolumn{2}{c}{\textbf{A.\ Comp.}} &%
                \multicolumn{2}{c}{\textbf{PubMed}} &%
                \multicolumn{2}{c}{\emph{Avg.\ Rank}} \\%
                \& Shift & Quantifier &%
                AE & RAE & 
                AE & RAE & 
                AE & RAE & 
                AE & RAE & 
                AE & RAE & 
                AE & RAE 
                \\\toprule%
            },
        before reading={\setlength{\tabcolsep}{3pt}},
        table foot=\bottomrule,
        head to column names,
        late after line={\ifthenelse{\equal{\approach}{CC}\or\equal{\approach}{PCC}\or\equal{\approach}{MLPE}}{\ifthenelse{\equal{\model}{MLP}\and\equal{\approach}{PCC}}{\\\midrule[\heavyrulewidth]}{\\\midrule}}{\\}},
        filter={\not\equal{\approach}{MLPE}},
    ]{#1}{}{
        \ifthenelse{\equal{\approach}{CC}\or\equal{\approach}{PCC}}{\multirow{\groupSize}{*}[0em]{\shortstack{\textbf{\model}\\\modelSuffix}}}{\ifthenelse{\equal{\approach}{MLPE}}{\modelSuffix}} & \approach &%
        \tblres{\coraMlAe}{\coraMlAeSe}{\coraMlAeBest} & \tblres{\coraMlRae}{\coraMlRaeSe}{\coraMlRaeBest} & 
        \tblres{\citeSeerAe}{\citeSeerAeSe}{\citeSeerAeBest} & \tblres{\citeSeerRae}{\citeSeerRaeSe}{\citeSeerRaeBest} & 
        \tblres{\photosAe}{\photosAeSe}{\photosAeBest} & \tblres{\photosRae}{\photosRaeSe}{\photosRaeBest} & 
        \tblres{\computersAe}{\computersAeSe}{\computersAeBest} & \tblres{\computersRae}{\computersRaeSe}{\computersRaeBest} & 
        \tblres{\pubMedAe}{\pubMedAeSe}{\pubMedAeBest} & \tblres{\pubMedRae}{\pubMedRaeSe}{\pubMedRaeBest} & 
        \tblres{\aeRank}{}{\aeRankBest} & \tblres{\raeRank}{}{\raeRankBest} 
    }}%
\begin{table*}[p]
    \caption{Quantification results (absolute error and relative absolute error).}\label{tbl:results}
    \centering
    \quantTable{tables/results.csv}
\end{table*}
\Cref{tbl:results} shows the mean quantification performance for all combinations of quantifiers, classifiers, distributions shifts and datasets.
Additionally, the last block of columns shows the average rank of each quantifier across all datasets for all combinations of classifiers and distribution shifts.
\textbf{Bold numbers} indicate that there is no statistically significant difference between the reported mean and the best mean within a given block, determined by the 95th percentile of a one-sided t-test.
The \ac{ppr} quantifiers use \ac{sis} with the interpolated \ac{ppr} kernel $\kappa_\lambda$ for different values of $\lambda$.

Overall, looking at the average ranks, we find that KDEy with \ac{sis} outperforms KDEy without \ac{sis} and, both \ac{pcc} and \ac{pacc}.
The results are consistent across all three types of distribution shift, all model types and, both the \ac{ae} and \ac{rae} metrics.
Under \ac{pps}, where \ac{sis} is not necessary, \ac{sis} generally does not significantly improve the quantification performance; nonetheless, we note that KDEy with \ac{sis} has a better average rank than KDEy without \ac{sis}.

\paragraph{Influence of the Classifier}
Unsurprisingly, the choice of classifier has a significant impact on the quantification performance.
Even though, a good classifier $h$ is not required by \ac{ql} to obtain an unbiased estimate of the label prevalences, the quality of this estimate is still correlated with the classifier's accuracy.
Overall, the structure-unaware \ac{mlp} classifier thus performs worst while \ac{appnp} performs best.

\paragraph{Influence of the Type of Covariate Shift}
The $\kappa_\lambda$ used in our experiments is based on the assumption that the distribution shift is induced by sampling localized \aclp{rw}.
In the \ac{rw} covariate setting, this assumption is satisfied, while in the \ac{bfs} setting, the \ac{ppr} kernel is, at least in theory, not appropriate.
Since the test vertices are selected by \ac{bfs} based on their distance to some start vertex, a \acs{sp}-based kernel, as in \cref{eq:sp-kernel}, seems plausible here.
However, our results show that a perfect match between the \ac{sis} kernel and the underlying distribution shift is not necessary.
In fact, the \ac{ppr} kernel performs well even in the \ac{bfs} setting, clearly outperforming all other quantifiers, while the \ac{sp} kernel performs comparatively poorly.

\begin{figure}[t]
    \centering
    \includegraphics[width=0.49\linewidth]{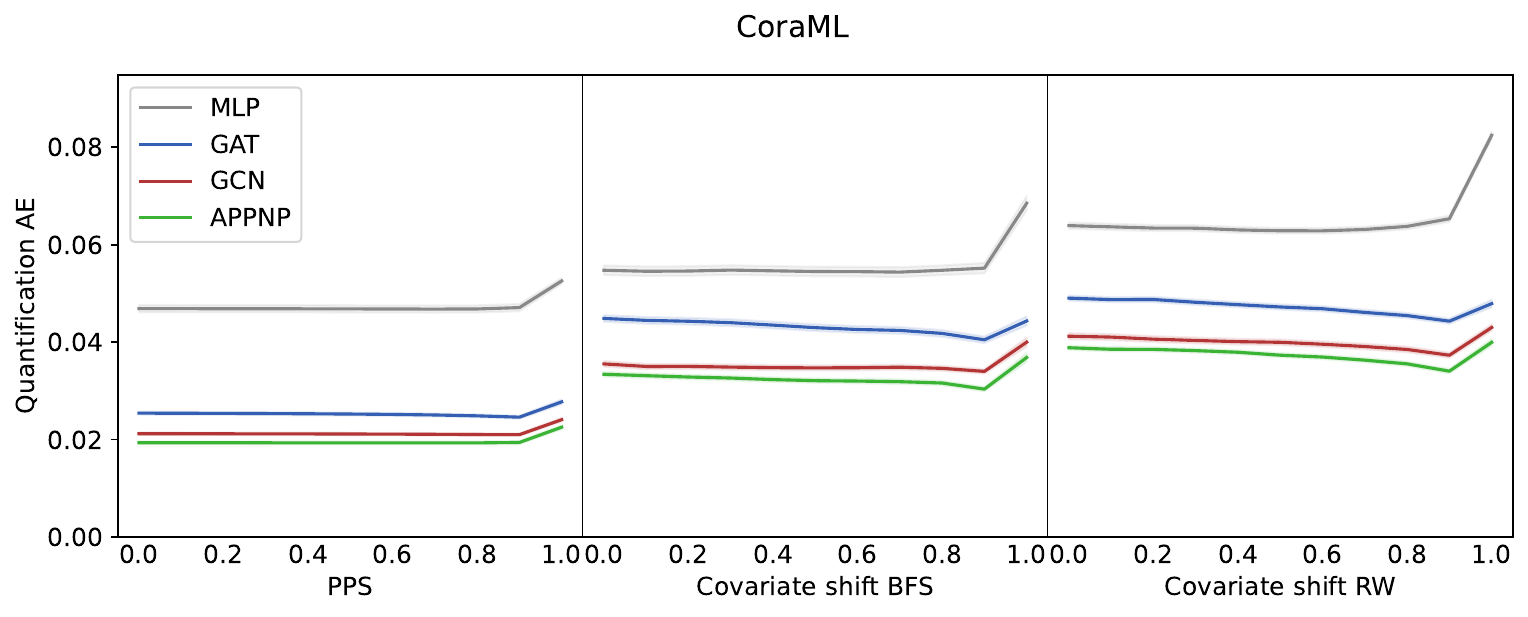}
    \includegraphics[width=0.49\linewidth]{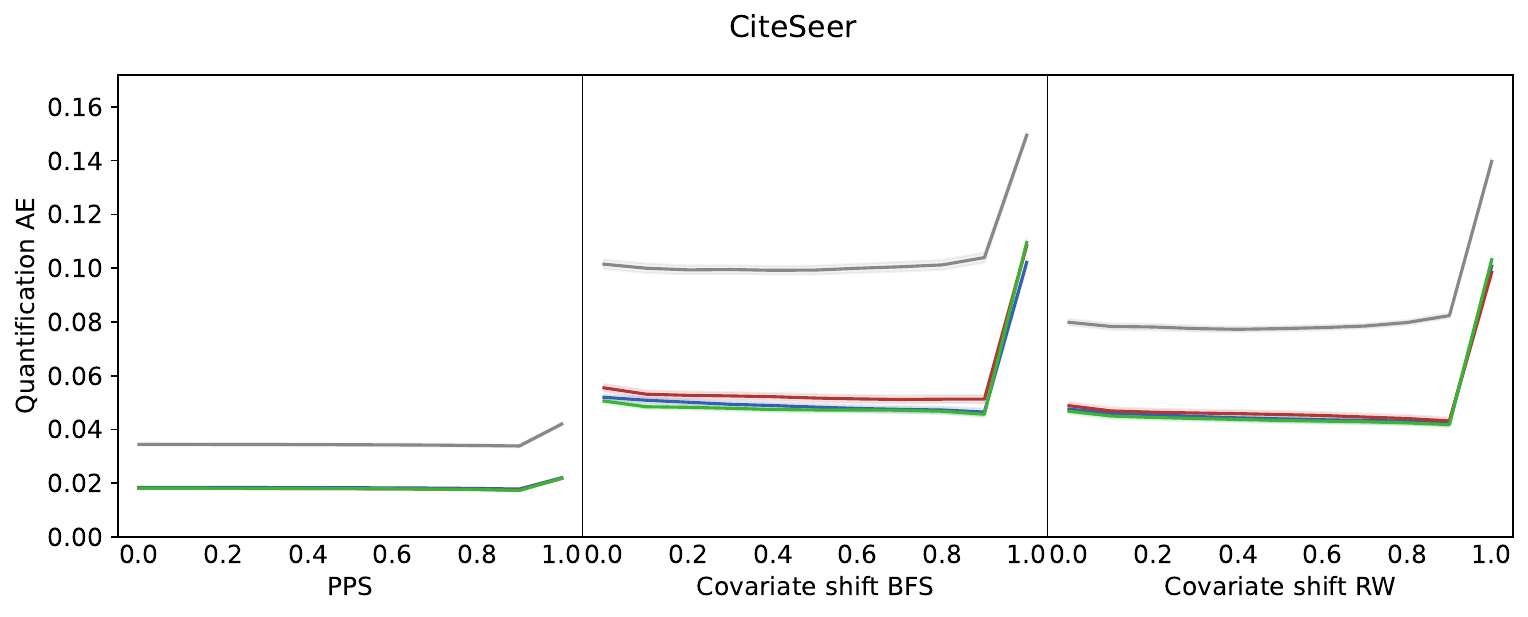} \\
    \includegraphics[width=0.49\linewidth]{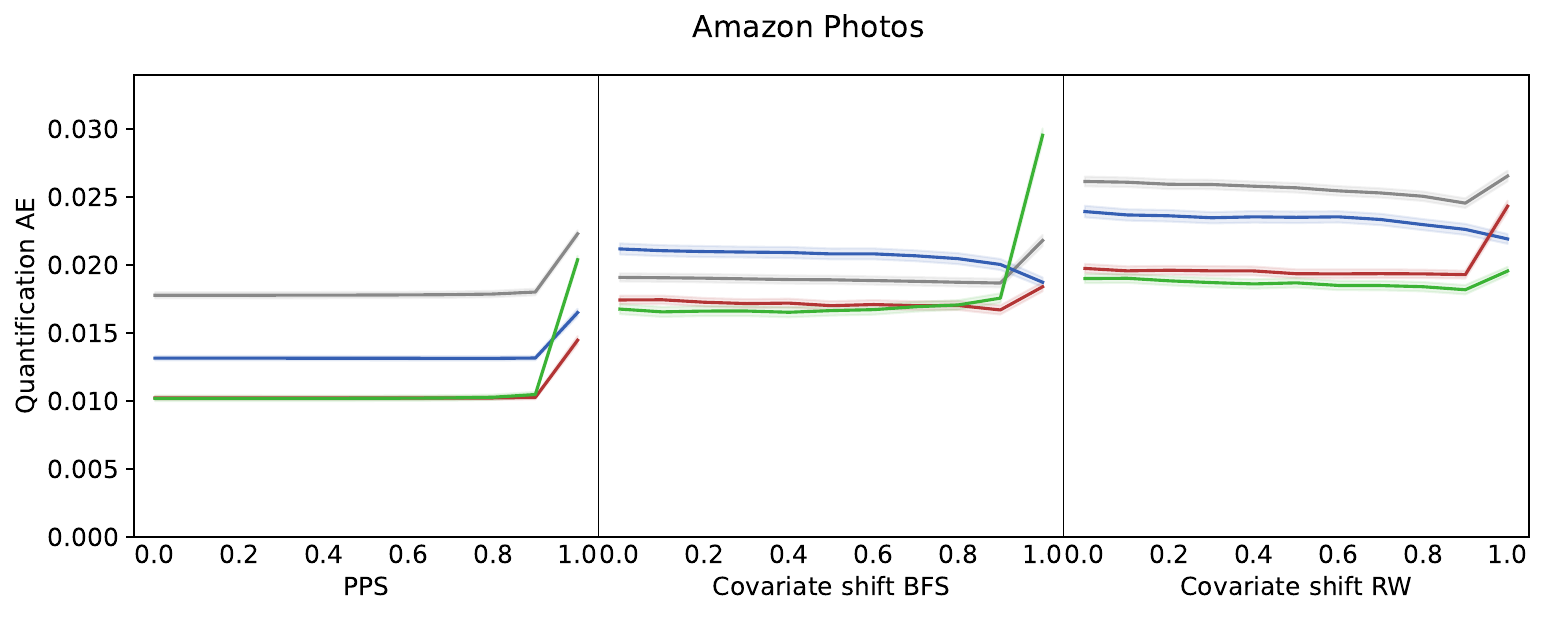}
    \includegraphics[width=0.49\linewidth]{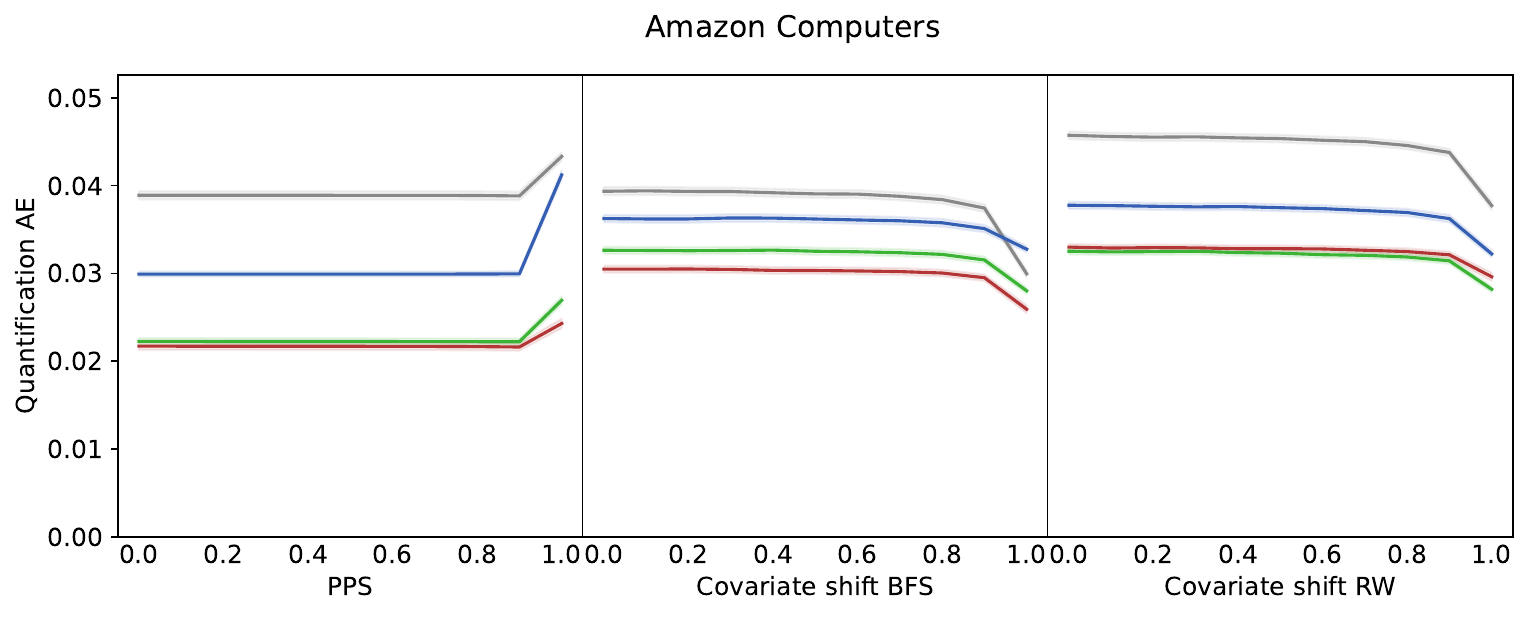}
    \caption{Quantification performance of KDEy with \ac{sis}, using the \ac{ppr} kernel $\kappa_\lambda$ for different values of $\lambda$.}\label{fig:ppr-int}
\end{figure}

\paragraph{Influence of $\kappa$ on SIS}
Note that the performance of \ac{sis} strongly depends on the choice of the kernel $\kappa$.
For $\lambda = 1$, the \ac{ppr} kernel performs poorly on all datasets except Amazon Computers.
\Cref{fig:ppr-int} shows that decreasing $\lambda$ slightly to $0.9$ already improves the performance significantly, further decreasing $\lambda$ then has little to no effect.
This illustrates an important tradeoff to consider when using \ac{sis}:
By making $\kappa$ more aggressive, in the sense that little to no weight is assigned to distant vertices, one can in-principle improve the performance of \ac{sis} by reducing the influence of irrelevant or misleading vertices from different regions of the graph.
However, if too many vertices are excluded, the effective sample size for the estimate $\hat{q}(\hat{s} \mid y)$ is reduced, making it more noisy.

The dataset-dependent optimal $\lambda$ value differences can be explained by different connectivity patterns in the datasets.
For example, while the CiteSeer dataset consists of multiple disconnected components, the Amazon Computers dataset mostly consists of a single large connected component (excluding a few disconnected outlier vertices).
If all vertices in a structurally shifted test set are sampled from a single (small) connected component, the \ac{ppr} kernel with $\lambda = 1$ will assign zero weight to all training vertices that are not in the same component, resulting in noisy estimates based on only a few vertices.
For less connected datasets, where sampling a test set from a small component is more likely, it is therefore often beneficial to assign at least some weight even to disconnected vertices, which is achieved by using a $\lambda < 1$.
For a well-connected dataset, such as Amazon Computers, this is not necessary.

To summarize, we have seen that KDEy combined with \ac{sis} and the \ac{ppr} kernel perform very well across different datasets and shift types, corroborating that \ac{sis} is able to effectively account for (structural) covariate shift given an appropriate kernel.

\section{Conclusion}\label{sec:conclusion}

We proposed a novel approach to quantification under structural covariate shift extending \ac{sis} from \ac{acc} to the KDEy quantification method.
We showed the effectiveness of this approach on a set of benchmark datasets with different types of distribution shift.
For future work, it would be interesting to investigate whether \ac{sis} can also be applied outside of the graph domain, e.g., in the context of timeseries data or geospatial data, where covariate shifts might occur in time or space.
Second, a more thorough analysis of the influence of the choice of the kernel $\kappa$ on the quantification performance is needed, especially since the choice of $\kappa$ is crucial for the performance of \ac{sis}.
Third, in this work we focused on the combination of \ac{sis} and KDEy, since KDEy is a state-of-the-art quantification method within the \ac{dm} framework.
Making other \ac{ql} methods that assume \ac{pps} applicable to covariate shifts would be another avenue for future work.
More specifically, investigating the combination of non-aggregative quantification approaches~\citep{qi2020} would be interesting.

\begin{credits}

    \subsubsection{\discintname}
    The authors have no competing interests to declare that are
    relevant to the content of this article.
\end{credits}
\end{document}